# A Comparative Study of Filters and Deep Learning Models to predict Diabetic Retinopathy

Roshan Vasu Muddaluru
Department of Computer Science & Engineering
Amrita School of Computing, Bengaluru
Amrita Vishwa Vidyapeetham, India
roshanvasu7@gmail.com

Sharvaani Ravikumar Thoguluva
Department of Computer Science & Engineering
Amrita School of Computing, Bengaluru
Amrita Vishwa Vidyapeetham, India
sharvaanit@gmail.com

Shruti Prabha
Department of Computer Science & Engineering
Amrita School of Computing, Bengaluru
Amrita Vishwa Vidyapeetham, India
shruti.prabha001@gmail.com

Tanuja Konda Reddy
Department of Computer Science & Engineering
Amrita School of Computing, Bengaluru
Amrita Vishwa Vidyapeetham, India
k.tanujareddy@gmail.com

Suja Palaniswamy*
Department of Computer Science & Engineering
Amrita School of Computing, Bengaluru
Amrita Vishwa Vidyapeetham, India
p_suja@blr.amrita.edu

*Abstract*— The retina is an essential component of the visual system, and maintaining eyesight depends on the timely and accurate detection of disorders. The early-stage detection and severity classification of diabetic retinopathy (DR), a significant risk to the public's health is the primary goal of this work. This study compares the outcomes of various deep learning models, including InceptionNetV3, DenseNet121, and other CNN-based models, utilizing a variety of image filters, including Gaussian, grayscale, and Gabor. These models could detect subtle pathological alterations and use that information to estimate the risk of retinal illnesses. The objective is to improve the diagnostic processes for DR, the primary cause of diabetes-related blindness, by utilizing deep learning models. A comparative analysis between Greyscale, Gaussian and Gabor filters has been provided after applying these filters on the retinal images. The Gaussian filter has been identified as the most promising filter by resulting in 96% accuracy using InceptionNetV3.

## I. INTRODUCTION

DR, an inflammatory condition affecting the retina (the light-sensitive tissue in the back of the eye), is one of the most prevalent and potentially blinding complications of diabetes mellitus. Degenerative conditions such as diabetic retinal disease are characterized by vascular injury to the retina, primarily brought on by prolonged high blood sugar levels. More than one-third of individuals with diabetes experience some kind of diabetic retinopathy; if not treated suitably, this condition can worsen from moderate non-proliferative abnormalities to proliferative diabetic retinopathy (PDR), which can be blindness, and diabetic macular edema (DME). By 2030, over 11 million people worldwide are predicted to suffer from advanced DR due to the rising incidence of diabetes. When aberrant blood vessels appear in DR, it is an indication of a complex pathological process that progresses through various phases. These phases range from moderate non-proliferative modifications to more severe and advanced proliferative modifications. Ensuring an early diagnosis and accurately classifying the severity of DR are critical goals. Accurate assessments of this kind are essential to enable effective therapeutic interventions and putting preventative measures in place to avoid irreversible sight loss.

The convergence of medical science and artificial intelligence has resulted in a paradigm shift in the diagnosis and management of DR in recent years. Traditional procedures frequently struggle to provide complex severity classifications, and more advanced methodologies are desperately needed. This study aims to bridge the gap by introducing a novel categorization approach that overcomes the constraints of existing automated grading systems. In this work, the potential of deep learning, specifically convolutional neural networks (CNNs), is leveraged to offer a robust and efficient approach for distinguishing diabetic retinopathy severity stages. This method not only offers improved accuracy, but it also aims to address the clinical care spectrum, which includes the five expert-defined severity stages: no apparent retinopathy, mild, moderate, severe non-proliferative diabetic retinopathy (NPDR), and proliferative diabetic retinopathy (PDR). The goal of this study is to delve into the intricate aspects of retinal images and make use of the capabilities of modern technology to contribute to the refining of diagnostic techniques and, in the end improve the standard of care for patients with DR.

The advancement of deep learning technology, particularly CNNs, has made it feasible to automate the analysis of retinal imagery. This has made the process of detecting DR more accurate and efficient.

The main goal of this project is to develop a methodology for categorizing retinal images that will be more effective at discriminating between the five different severity classifications of DR than the automated grading systems that are currently in use. Currently, most published models simplify the disease category for binary identification of diabetic retinopathy without considering the degree of development. Although some studies classified three grades, few studies distinguish between the five expert-defined severity stages that directly affect clinical care.

This paper's novelty is found in the methodical research and assessment of deep learning models with various filters, with the goal of utilizing the best filters for maximum efficiency. A high accuracy of 96% is achieved by InceptionNetV3, by using the Gaussian filter.

The remaining sections of the paper is organized as follows: Section II describes the literature survey. Section III introduces the data and how it was gathered. Section IV discusses the proposed methodology. This is followed by discussion of the results and their analysis and lastly, Section VI summarizes about the conclusion and future scope.

## II. LITERATURE SURVEY

A deep learning framework that can accurately identify a number of common fundus illnesses and disorders is presented by the authors in [1]. The methodology entailed building a two-level hierarchical system for the categorization of 39 different types of diseases and conditions using three sets of convolutional neural networks and a Mask-RCNN. For multi-label classification in the main test dataset, the findings revealed an AUC of 0.9984, sensitivity of 0.978, specificity of 0.996, and frequency-weighted average F1 score of 0.923.

In order to tackle the multi-label long-tailed issue, the work done in [2] employs hybrid knowledge distillation and an instance-wise class balanced sampling technique. To train retinal illness recognition algorithms for the first time, the authors employed two internal and two external datasets, one of which included over a million fundus pictures covering over 50 retinal diseases. The outcomes in [3] demonstrate the benefits of deeper residual networks over their simpler equivalents, solving the degradation issue and reaching cutting-edge functionality. The investigation explores how response strength and signal modification are affected by network depth, offering insights into residual function behavior. In order to identify diabetic retinopathy, the work done in [4] uses a methodology that includes picture preprocessing, feature extraction, and classification using SVM. The results reveal that retinal abnormalities can be detected with great sensitivity and accuracy, ranging from 84.31% to 94%. The data highlights how important automated screening systems are for accurately identifying retinal hemorrhages, which helps with early diagnosis and prevents vision loss.

Morphological operations and machine learning approaches are used in the [5] described methodology for the early diagnosis of diabetic retinopathy. The findings indicate that in grading the severity of retinal pictures, the SVM classifier performs better than the KNN classifier. The approach described in [6] outperforms previous methods in datasets related to malaria and diabetic retinopathy by using intensity and compactness features in a multi-level superpixel architecture to detect retinal leakage. Its potential for clinical applications is indicated by the results, which show great sensitivity and specificity, comparable to human specialists, and accurate segmentation of leaking zones.

Morphological operations and segmentation techniques are used in the work of [7] to detect microaneurysms, exudates, and blood vessels. A variety of features are retrieved from the retinal fundus image, and these features are then subjected to Haar wavelet modifications. Subsequently, a single rule classifier and a backpropagation neural network are employed to identify the images as indicative of diabetes or not. In order to facilitate additional processing, the blood vessels and optic disk are extracted by the authors of [8], who also eliminated the background pixels. After that, they used a filter bank to remove any potential lesion areas. Each region's features are then taken out in order to aid in classification. To increase classification accuracy, a hybrid classifier that combines an m-Mediods classifier and a Gaussian mixture model is presented. The classifier is optimized by genetic algorithms and cross-validation. A back propagation neural network was used in [9] to assess fundus photographs of 147 diabetic patients and 32 healthy individuals. Vascular, exudative, and hemorrhagic traits were among those for which the network was trained. The network successfully detected 93.1% of exudates, 73.8% of hemorrhages, and 91.7% of arteries. According to the researchers' findings, the neural network effectively identified arteries, exudates, and hemorrhages in the fundus images.

Due to its effectiveness, K-nearest neighbors were utilized for red lesions and Gaussian mixture models for bright lesions in [10]. Classification speed is improved by 94-97% while applying AdaBoost to reduce features from 78 to 30. The two-step hierarchical categorization is much faster than the latter. The DR severity grade is determined in the third stage by combining the lesion information. The deep learning model presented by the authors in [11] was found to perform better, and regression activation maps (RAM) can offer interpretability and insight into the algorithm's decision-making process. The suggested convolutional network provides the RAM for visual explanation and achieves competitive performance when compared to state-of-the-art techniques, according to experiments conducted on a sizable dataset.

Yang in [12] presented a two-phase deep convolutional neural network method for assessing the severity of diabetic retinopathy and identifying lesions. In fundus image patches, the first local network finds lesions such as microaneurysms, hemorrhages, and exudates. The second global network uses a weighted lesion map created from the local network outputs to rank the severity of the entire fundus image. Patches with more severe lesions are given greater weight on this weighted lesion map, which enhances grading performance.

According to the researchers in [13], training wide residual Inception networks requires appropriate activation scaling in order to stabilize the process. When combined, the four models produce state-of-the-art results on the ImageNet classification challenge, illustrating that bigger model sizes and residual connections can both enhance picture recognition ability. Pre-processing methods like erosion were employed by the authors in [14] to eliminate insignificant features from the retinal periphery and histogram equalization to improve contrast. Compared to using the original images without pre-processing, the results demonstrate that the pre-processed images increase the accuracy of the CNN model as well as other pre-trained models like AlexNet, VGG16, and ResNet50. The pre-processing of the photos increased the accuracy of the suggested CNN model from 80% to 86%.

A CNN approach is proposed by the study in [15] to identify diabetic retinopathy in fundus images with 92% accuracy. Grad-CAM, an explainable AI approach, is used to improve the model by visualizing the regions that the model concentrates on in order to provide predictions. In [16], the authors used deep learning to segment and resize the photos. The photos' texture features were then extracted using the Gabor filter. In [17], images were denoised using four distinct filters and in [18], a deep learning ensemble network was employed.

The primary aim of this research is to conduct a comparative analysis of the filters applied to the dataset with a focus on identifying and assessing those that yield optimal results. Through a systematic examination of various filters, the study aims to discern and highlight the most effective ones in enhancing the dataset. This investigation is geared towards advancing our understanding of the impact of different filtering techniques, ultimately contributing valuable insights to the field and informing best practices in data processing.

## III. DATA DESCRIPTION

The greyscale DR dataset [19] includes 3664 retinal images that highlight different phases of diabetic retinopathy, a consequence of diabetes that is the main cause of visual impairment. The images show the evolution of an illness and are divided into five graded classes namely, Mild, Moderate, No_DR, Proliferate_DR and Severe which were labelled by medical specialists. The images are downsized to 224 × 224 pixels, which are the suitable dimensions for feeding them into deep learning neural networks. As seen in Fig 1., images having no sign of disease consume most of the dataset, and therefore the dataset is imbalanced. Therefore, augmentation was done on the images. The applied image augmentation strategy encompasses pixel normalization for values within [0, 1], coupled with random horizontal and vertical shifts of up to 20% during training. This intentional perturbation introduces spatial diversity, simulating real-world variations. The strategy aims to mitigate overfitting by exposing the model to a spectrum of transformed images, fostering better generalization. Additionally, 20% of the dataset is allocated to a validation subset for robust model evaluation.

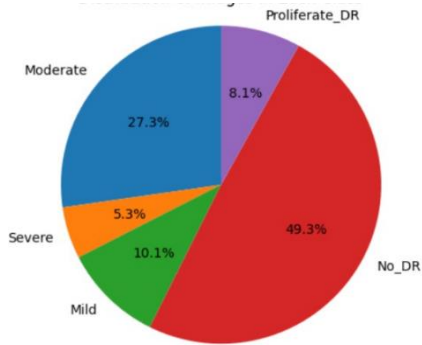

Fig 1. Original Distribution of Images

After augmentation, three distinct filters were applied, namely Gaussian, Grayscale, and Gabor, and then their impact was evaluated using various models. The grayscale and gaussian images were already available in the dataset.

### A. Gaussian

The Gaussian filter shown in Fig. 2. (i), effectively smooths irregularities and reduces noise, particularly enhancing structures relevant to diabetic retinopathy such as blood vessels and lesions. By prioritizing pixel weights based on proximity, the filter contributes to the extraction of meaningful features crucial for classification. The preprocessed images then undergo feature extraction, capturing distinctive patterns indicative of diabetic retinopathy.

$$G(x,y) = \frac{1}{2\pi \sigma^2} \exp\left(-\frac{x^2+y^2}{2\sigma^2}\right) \qquad \text{Eq. (1)}$$

G(x) = Value of the Gaussian function at position (x,y) in the image
σ=8 (standard deviation of the Gaussian envelope)

### B. Greyscale

The grayscale filter on the eye shown in Fig. 2 (ii) proves advantageous for analyzing retinal eye images by simplifying data and enhancing contrast. This conversion to a single intensity channel streamlines image processing, making it compatible with various analysis techniques and reducing computational complexity. The resulting grayscale images offer improved visibility of critical anatomical structures, such as blood vessels, lesions, or abnormalities. The pixel intensities range from 0 to 255, 0 being black and 255 being white.

### C. Gabor Filter

The Gabor filter shown in Fig. 2. (iii) is distinguished by its exceptional capacity to capture complex texture patterns in a variety of orientations and scales. In contrast to normal filters, the Gabor filter excels at picking up on minute characteristics related to diabetic retinopathy, like changes in blood vessel and microaneurysms. This is because it mixes a sinusoidal function with a Gaussian envelope. Because of its unique feature extraction capacity, the Gabor filter is very useful for bringing attention to subtle textures that are suggestive of the illness. In contrast to traditional filters, the Gabor filter improves the accuracy of diabetic retinopathy diagnosis in retinal pictures by highlighting these particular patterns, which leads to a more sophisticated and sensitive categorization technique.

$$G(x,y;\lambda,\theta,\psi,\sigma,\gamma) = \exp\left(-\frac{x'^2+\gamma^2 y'^2}{2\sigma^2}\right)\cos\left(2\pi\frac{x'}{\lambda}+\psi\right) \qquad \text{Eq. (2)}$$

G(x,y) = filter response at position (x,y)
x' = xcos(θ)+ysin(θ)
y' = -xsin(θ)+ycos(θ)
λ=10 (wavelength of sinusoidal factor)
θ=45 degrees (orientation of the filter)
σ=8 (standard deviation of the Gaussian envelope)
γ=0.5 (spatial aspect ratio, which stretches the filter along one direction)
ψ=2 (Phase offset of sinuoisoidal factor)

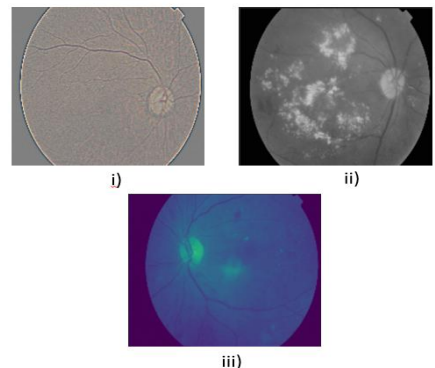

Fig 2. Filters applied on the images i) Gaussian ii) Grayscale iii) Gabor

### D. System Architecture

The process of analyzing retinal pictures with deep learning models and different filters is depicted in the figure Fig. 3. Retinal images are used as the input and these images are resized to 224x224 and augmented. The augmented photos are then subjected to three filters: the Gaussian, grayscale, and gabor filters. With the aid of these filters, particular picture characteristics that are pertinent to the categorization of diabetic retinopathy can be extracted.

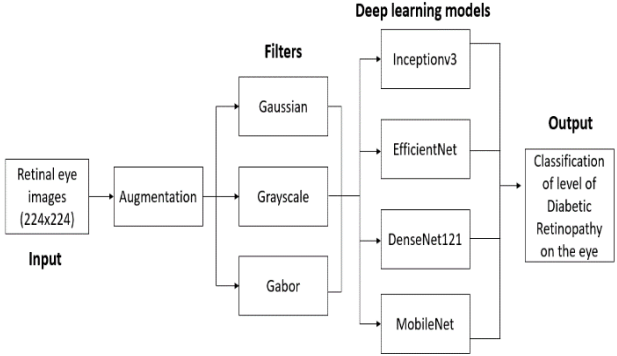

Fig. 3. Proposed System Architecture

The four deep learning models InceptionNetV3, EfficientNet, DenseNet, and MobileNet are fed the pre-processed images. These models are intended to evaluate the information retrieved by the filters and forecast whether or not diabetic retinopathy will be present in the retinal images, as well as how severe it will be.

## IV. PROPOSED METHODOLOGY

The outcomes produced by various deep learning models utilizing various filters—the gaussian, grayscale, and gabor filters are compared in this paper.

### A. EfficientNetB0

Compound scaling is used by EfficientNetB0 to systematically balance the trade-offs between computing efficiency and model complexity, starting with a baseline network. In order to improve feature representation, it presents a novel mobile inverted bottleneck block that combines linear bottleneck layers and inverted residuals. In addition to achieving state-of-the-art performance across a variety of image recognition tasks, EfficientNetB0 with its inventive building blocks and efficient scaling proves advantageous for extracting meaningful features from retinal images, where computational efficiency is critical for practical applications.

### B. InceptionNetV3

Factored convolutions are a feature of InceptionNetV3's architecture shown in Fig 4. and they are known to lower computational burden and redundancy while increasing efficiency. Auxiliary classifiers in the network facilitate gradient flow during training, which leads to a more stable convergence. InceptionNetV3, which has roughly 23 million parameters, is a reasonably moderate number that balances model complexity and resource efficiency. This makes it an excellent choice for feature extraction from retinal images, where minute details and contextual information are crucial for precise medical diagnosis.

### C. DenseNet121

With its tightly connected blocks, DenseNet121's architecture encourages feature reuse while improving information flow between layers and gradient flow. In addition to allowing the model to take advantage of a rich feature hierarchy, this dense connection lowers the possibility of vanishing gradients. DenseNet121 is useful for retinal image analysis because it can effectively capture global and local features that are important for recognizing minute details that may indicate different visual diseases. The model is well-suited for resource-constrained contexts in medical imaging applications because of its moderate parameter count of about 8 million, which enhances computational efficiency without sacrificing the model's capacity to learn complex representations.

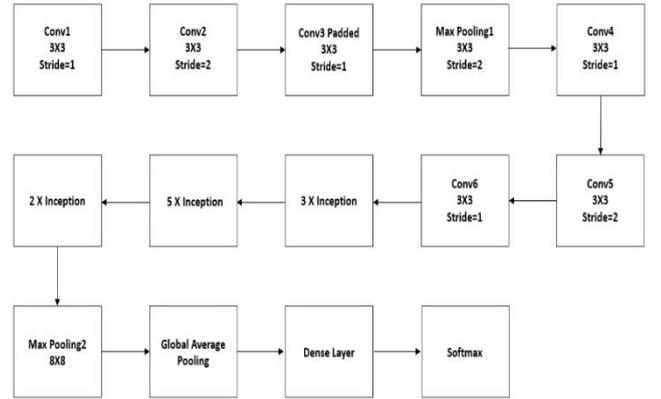

Fig. 4. Proposed architecture of InceptionNetV3

### D. MobileNetV2

A thin neural network architecture called MobileNetV2 is intended for mobile and edge devices. Inverted residuals and linear bottlenecks are utilized to effectively capture hierarchical characteristics. Its compact architecture, with about 3.4 million parameters, offers a reasonable compromise between computing efficiency and feature representation, making it appropriate for resource-constrained applications such as retinal image processing. The architecture is well-suited for extracting pertinent information from retinal images for tasks like disease diagnosis or anomaly detection because of its depth wise separable convolutions, which speed up computation while maintaining important features.

## IV. RESULTS AND ANALYSIS

The performance of multiple deep learning models on the task of diabetic retinopathy severity classification using retinal pictures with different filters are shown in Table 1. EfficientNetB0, InceptionNetV3, DenseNet121, and MobileNetV2 are the models under consideration. The filtered datasets were used to train the deep learning models, and test accuracy was used to determine benchmarks. Accuracy was tested independently on the training and test sets for each model-filter combination over numerous trials.

The results in Table 1. shows the accuracies of the mode, one for each dataset-trial combination.Table 2. displays the hyperparameters that were used for training the model. Adam is effective due to its adaptive learning rates, momentum for faster convergence, and efficient memory usage. ReLU activation functions can improve feature learning in convolutional neural networks by encouraging non-linearity, which helps the model capture complex patterns essential for differentiating between phases of diabetic retinopathy and improve the severity classification of the condition.

Table 1. Comparison of Results

| Model | **Greyscale** | | **Gaussian** | | **Gabor** | |
|---|---|---|---|---|---|---|
| | Train | Test | Train | Test | Train | Test |
| EfficientNetB0 | 0.908 | 0.632 | 0.87 | 0.74 | 0.97 | 0.69 |
| **InceptionNetV3** | **0.98** | **0.92** | **0.98** | **0.96** | **0.86** | **0.85** |
| DenseNet121 | 0.82 | 0.71 | 0.74 | 0.76 | 0.73 | 0.73 |
| MobileNetV2 | 0.81 | 0.65 | 0.73 | 0.68 | 0.78 | 0.70 |

Table 2. Parameters used by the models

| Epochs | Optimizer | Activation Function | Batch Size |
|---|---|---|---|
| 50 | Adam | ReLu | 32 |

Across trials, InceptionNetV3 outperforms various other models on the Gaussian filtered data, obtaining over 96% test accuracy. It is observed that comparatively, almost all the models reached their peak when trained with Gaussian images. InceptionNetV3 uses multiple convolutional layers with differing filter sizes to extract features at various scales. Its complex design with the usage of inception modules, which comprise numerous parallel convolutional procedures. This helps to capture both fine and coarse details in retinal pictures.

Gaussian filter is the best suited when it comes to classification and prediction of DR. One of the main reasons for this is, Gaussian filters are extensively employed in image noise reduction. Since these retinal images contain noise due to lighting conditions or poor-quality photography, the Gaussian filter actively smoothens it out and improves the overall image quality. The features enhanced by Gaussian filtering better reflect the particular characteristics of diabetic retinopathy. It was observed that the nerves and nerve endings were better highlighted in the images after applying this filter. These extracted features are more significant for differentiating between the distinct classes. Gabor filter mainly concentrated on contrasting regions on the retina but this did not help the models' effectively capture features. Greyscale images lacked clarity and did not provide good results.

The difference was significant when compared to the other accuracy range of 70 to 80 for other models. Its superiority, however, is less obvious on Greyscale filtered data, where accuracies are more tightly grouped. InceptionNetV3 and Densnet121 have reasonably stable filter performance, sitting in the 75-80% ranges. MobileNetV2 and EfficientNetB0, on the other hand, lag with test accuracy ranging from mid-60% to low-70%. It is normal for training and validation losses to first decline as the model learns.

It is common for the training loss to decrease over the course of the epochs as the model improves at predicting the training set. But after around five epochs, the validation loss starts to rise again after first declining. This is a sign of overfitting, a condition in which the model learns too much from the training set, including subtleties and noise that do not transfer to new data. Figure 5(i) shows overfitting due to the divergence of the training and validation curves. A noticeable divergence between the training and validation losses appears in Fig. 5(iii) after a few epochs, which often denotes overfitting. This suggests that training data-specific learning patterns can be accomplished using the model.

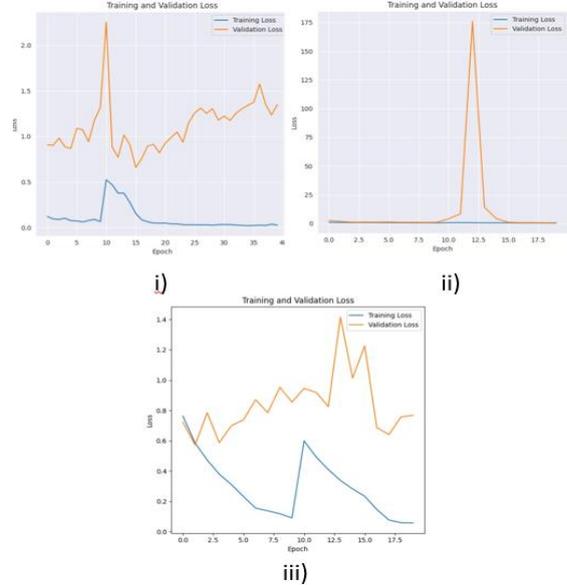

Fig 5. Loss Plots for Inception Net on
(i) Gaussian (ii) Greyscale (iii) Gabor

VI. CONCLUSION AND FUTURE SCOPE

The proposed work focuses on three different filters that were applied to the original greyscale images, the Gaussian filter emerged as the most promising filter among those that were used. This is due to the fact that features improved by Gaussian filtering in conjunction with the InceptionNetV3 model shown a greater ability to capture the crucial features of diabetic retinopathy. This filter improved the visibility of the blood vessels and nerves in the pictures. The significance of these extracted traits lies in their ability to distinguish between the various classes. The Gabor filter primarily focused on contrasting areas of the retina, although this was ineffective in assisting the models in successfully capturing information.

The current work could be expanded upon and improved by using different filters on the initial grey-scaled dataset. Furthermore, an increase in the visibility of blood vessels and nerves might stem from having access to a variety of pre-processing techniques like Region of Interest (ROI) trimming and histogram equalization. This could improve the accuracy of the model for the early detection of diabetic retinopathy. The Vision Transformer (ViT) is another significant model that might be applied to this dataset because it has been shown to perform better than more conventional deep learning models.